\documentclass[a4paper,conference,10pt]{IEEEtran}
\ifCLASSINFOpdf
\else
\fi
\ifCLASSOPTIONcompsoc
  \usepackage[caption=false,font=normalsize,labelfont=sf,textfont=sf]{subfig}
\else
  \usepackage[caption=false,font=footnotesize]{subfig}
\fi

\usepackage{epsfig}
\usepackage{graphics}
\usepackage{graphbox}
\usepackage{amsmath}
\usepackage{amssymb}

\usepackage{booktabs} 
\usepackage{placeins}
\usepackage{mathtools}
\usepackage{multirow}
\usepackage{algorithm, algpseudocode}
\usepackage{hyperref}

\DeclarePairedDelimiter\ceil{\lceil}{\rceil}

\DeclareMathOperator*{\diag}{diag}
\DeclareMathOperator*{\err}{err}
\DeclareMathOperator*{\cost}{cost}

\DeclareMathOperator*{\round}{round}
\DeclareMathOperator*{\resize}{resize}

\newcommand{\vpad}{0.05cm}
\newcommand{\imgwidth}{1.8cm}

\begin{document}
%
\title{Fast Multi-Level Foreground Estimation}

\author{\IEEEauthorblockN{Thomas Germer, Tobias Uelwer, Stefan Conrad, Stefan Harmeling}
\IEEEauthorblockA{Department of Computer Science\\
Heinrich-Heine-Universit\"at D\"usseldorf\\
Email: \{thomas.germer, tobias.uelwer, stefan.conrad, stefan.harmeling\}@hhu.de}}


%


\maketitle

\begin{abstract}
	Alpha matting aims to estimate the translucency of an object in a given image. The resulting alpha matte describes pixel-wise to what amount foreground and background colors contribute to the color of the composite image.
	While most methods in literature focus on estimating the alpha matte, the process of estimating the foreground colors given the input image and its alpha matte is often neglected, although foreground estimation is an essential part of many image editing workflows.
	In this work, we propose a novel method for foreground estimation given the alpha matte.
	We demonstrate that our fast multi-level approach yields results that are comparable with the state-of-the-art while outperforming those methods in computational runtime and memory usage.
\end{abstract}


%
\IEEEpeerreviewmaketitle

\section{Introduction}

For a color image $I$ with foreground pixels $F$ and background pixels $B$, the alpha matting problem asks to determine opacities $\alpha$, such that the equality
\begin{equation}
I = \alpha F + (1-\alpha)\label{eq:compositing}B
\end{equation}
holds. 
Equation \ref{eq:compositing} is called the compositing equation.
Alpha matting can be seen as an attempt to undo the compositing equation to get the original $\alpha$. 
In this work we want to focus on the problem of estimating the foreground pixels $F$ given the image $I$ and the matte $\alpha$.
A naive method to compose an image on a new background is to use $I$ in place of $F$, obtaining a new image $I^\text{new} = \alpha I + (1 - \alpha) B^\text{new}$, but this is only sufficient if $\alpha$ is close to binary, i.e. $\alpha$ is almost 0 or 1. 
This naive approach results in background colors that bleed through partially transparent regions, as visualized in Figure~\ref{fig:problem}~(c).

\newcommand{\problemImageWidth}{3.2cm}

\begin{figure}%
	\centering
	\subfloat[Input image]{{\includegraphics[width=\problemImageWidth, trim={0 8cm 0 0}, clip]{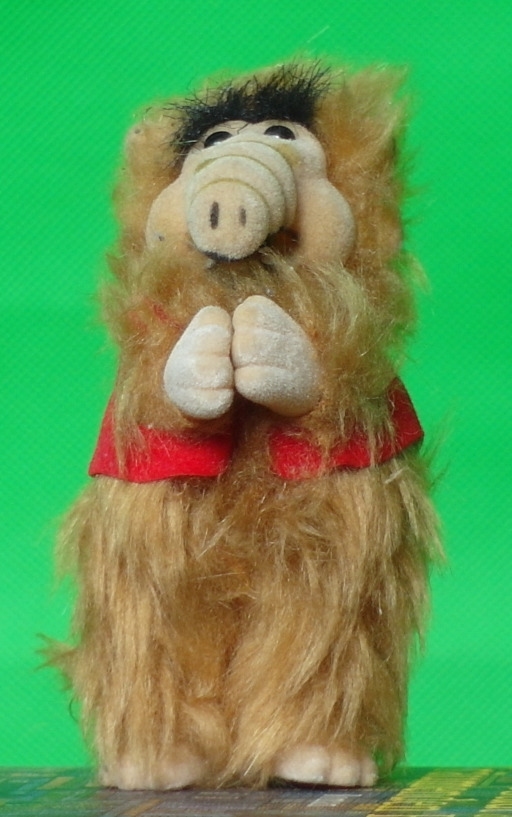} }}%
	\quad
	\subfloat[Input alpha matte]{{\includegraphics[width=\problemImageWidth, trim={0 8cm 0 0}, clip]{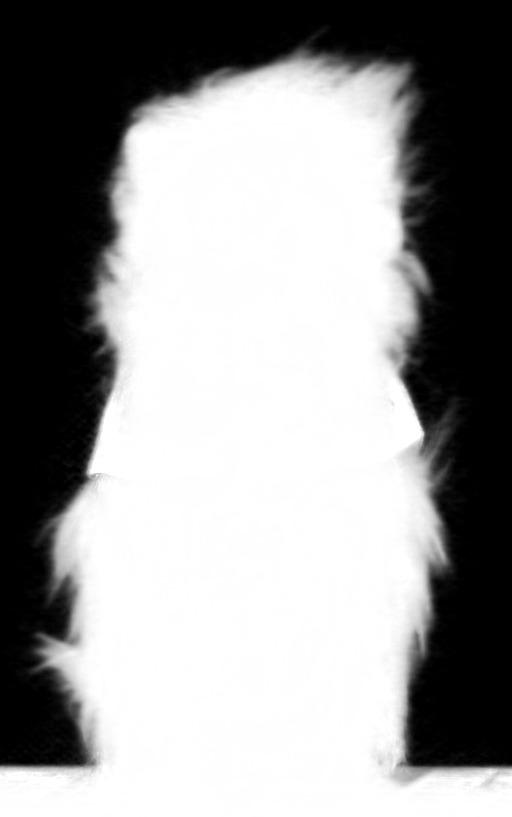} }}%
	\\
	\subfloat[Input image naively composed on white background]{{\includegraphics[width=\problemImageWidth, trim={0 8cm 0 0}, clip]{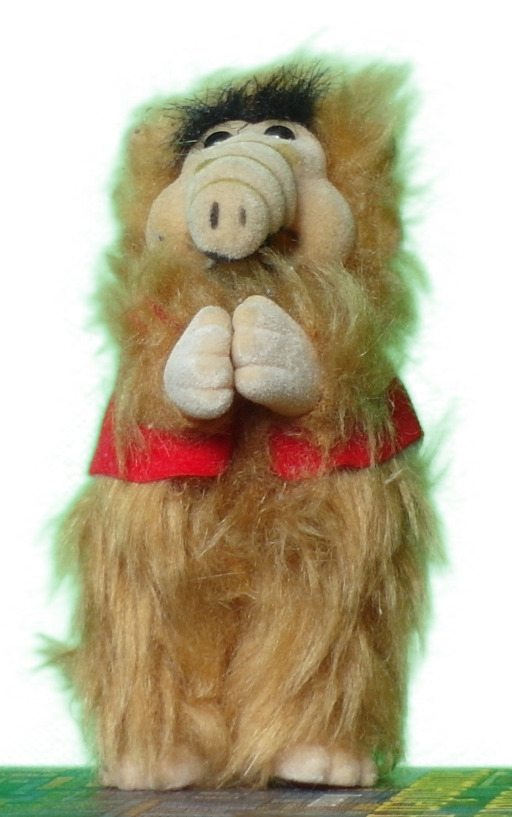} }}%
	\quad
	\subfloat[Computed foreground composed on white background]{{\includegraphics[width=\problemImageWidth, trim={0 8cm 0 0}, clip]{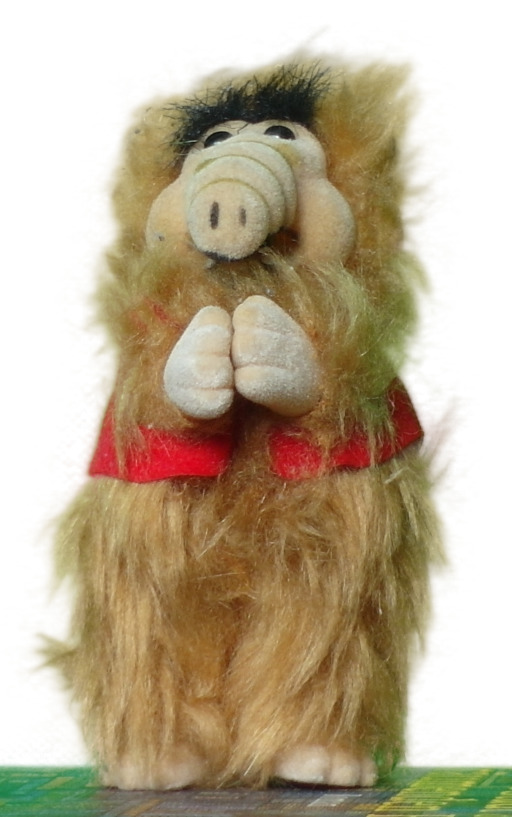} }}%
	\caption{Foreground estimation aims to compute the foreground given an input image (a) and input alpha matte (b). The obtained foreground can then be composed onto new a background (d) without the old background showing though (c).}
	\label{fig:problem}%
\end{figure}

\section{Related Work}
The problem of estimating the alpha matte is well-studied in literature. 
Recently, neural network based methods were introduced to estimate the alpha values. 
Cho et al.~\cite{cho2016natural} compute alpha in an end-to-end fashion based on outputs of other methods using a convolutional neural network.
Xu et al.~\cite{xu2017deep} train an encoder-decoder network to predict alpha and a refinement network to improve the prediction.
Lutz et al.~\cite{lutz2018alphagan} employ a generative adversarial network.
Cai et al.~\cite{cai2019disentangled} stack a recurrent neural network onto an autoencoder network to first estimate an optimal ternary segmentation followed by the alpha matte in a multi-task learning setting.

However, these methods were not devised to estimate the foreground colors.
In the following we want to give a brief overview over methods that are capable of foreground estimation.
Hou et al. \cite{hou2019context} train a network for local features and a network for global context information simultaneously. Tang et al. \cite{tang2019learning} train a chain of three neural networks to successively estimate background, foreground and alpha.
Levin et al. \cite{levin2007closed}, Chen et al. \cite{chen2013knn} and Aksoy et al. \cite{ifm} globally minimize a quadratic energy function based on a smoothness prior applied to foreground and background, k-nearest neighbors in color and pixel coordinate space and a combination thereof respectively.
All of those methods either have high memory requirements, high computation times, or both, which motivates the development of a faster method.

\section{Method}

\subsection{Notation}
We use $I_i^c$ to denote the intensity of color channel $c$ of image $I$ at index $i$. In addition, we make use of the notation $I_{i_x}$ by Levin et al. \cite{levin2007closed} to denote the gradient of the image towards the x-direction.
We use a similar notation for the foreground image $F$ and the background image $B$ respectively.

\subsection{Closed-Form Foreground Estimation}
In order to estimate both foreground and background images, Levin et al. \cite{levin2007closed} propose to minimize a cost function for each pixel $i$ and color channel $c$ consisting of three terms, one to constrain the resulting color from the compositing equation (Equation~\ref{eq:compositing}) and two to reduce the magnitude of color gradients $F_{i_x}, F_{i_y}, B_{i_x}$ and $B_{i_y}$ in regions of large $\alpha$-gradients $|\alpha_{i_x}|$ and $|\alpha_{i_y}|$, thereby preserving texture information
\begin{equation}
\begin{split} \label{eq:cfcolor}
\cost_{\text{global}}(F, B) = \sum_{i\in I} \sum_c &\left[ \alpha_i F_i^c + (1 - \alpha_i) B_i^c - I_i^c \vphantom{\left( B_{i_y}^c \right)^2} \right]^2\\
&+ |\alpha_{i_x}| \left[ \left( F_{i_x \vphantom{i_y}}^c \right)^2 + \left( B_{i_x \vphantom{i_y}}^c \right)^2 \right]\\
&+ |\alpha_{i_y}| \left[ \left( F_{i_y}^c \right)^2 + \left( B_{i_y}^c \right)^2 \right].
\end{split}
\end{equation}

We find that this closed-form color estimation method can be accelerated greatly with appropriate preconditioning, for example by employing a thresholded incomplete Cholesky decomposition in conjunction with conjugate gradient descent, but solving the resulting $2n$-by-$2n$ linear system still takes in the order of 30 seconds per color channel to converge below a residual error of $10^{-6}$ for an $n = 0.4$ megapixel image on current consumer hardware.
This is unsatisfactory for interactive image editing. Our goal for practical applications is a method which runs in a few seconds on multi-megapixel images on common hardware.

\subsection{Multi-Level Foreground Estimation}

A simplified approach might try to approximate the closed-form cost function by only solving it for a small local region instead of finding a global solution.
Unfortunately, this does not work because a local solution barely propagates foreground and background colors into the region with non-binary alpha values, even with many iterations.
However, a multi-level approach can alleviate this shortcoming, leading to an efficient method to approximate foreground and background colors.

To this end, we start with the cost function by \cite{levin2007closed}, which we modify for a local image region centered at the pixel $i$ for a fixed color channel $c$. The color gradients are expressed as a sum over the neighboring pixels $j \in N_i$. Furthermore, by adding a regularization factor $\epsilon_r$, we make the problem well-defined in regions with constant alpha values. Otherwise, foreground colors $F$ and background colors $B$ would be unconstrained in regions where the translucency $\alpha$ is 0 or 1 respectively. In addition, we introduce the constant $\omega$ to control the influence of the alpha gradient
\begin{equation}
\begin{split}
\cost_{\text{local}}(F_i^c, B_i^c) =
(\alpha_i F_i^c + (1 - \alpha_i)B_i^c - I_i^c)^2 \\
+ \sum_{j \in N_i} (\epsilon_r  + \omega |\alpha_i - \alpha_j|)\left[
(F_i^c - F_j^c)^2 +
(B_i^c - B_j^c)^2
\right].
\end{split}
\end{equation}

This cost function can be expressed in matrix form as
\begin{equation}
\begin{split}
\cost_{\text{local}}(g_i^c) =
\left(
U_i^T g_i^c
- I_i^c
\right)^2
+
(R g_i^c - h_i)^T V_i (R g_i^c - h_i)
\end{split}
\end{equation}
where $g_i^c = [F_i^c, B_i^c]^T$ is a vector of the foreground and background colors,
$U_i = [\alpha_i, 1 - \alpha_i]^T$ is a vector describing how to weight the colors and $h_i^c$ is a vector of the neighboring foreground and background colors
\begin{equation}h_i^c = 
\left[
\begin{array}{c}
\vdots\\
F_j^c\\
\vdots\\
\hline
\vdots\\
B_j^c\\
\vdots
\end{array}
\right],j \in N_i.
\end{equation}
Furthermore, $R$ is a $2|N_i|$-by-$2$ matrix to broadcast the local foreground and background colors to the size of vector $h_i$
\begin{equation}
R =
\left[
\begin{array}{c|c}
1_{|N_i|,1} & 0_{|N_i|,1} \\
\hline
0_{|N_i|,1} & 1_{|N_i|,1}
\end{array}
\right],
\end{equation}
and $V_i$ is a $2|N_i|$-by-$2|N_i|$ block matrix
\begin{equation}
V_i =
\left[
\begin{array}{c|c}
S_i & 0 \\
\hline
0 & S_i
\end{array}
\right].
\end{equation}
The  top-left and bottom-right blocks $S_i = \diag(s)$ of $V_i$ with entries $s_j = \epsilon_r + \omega |\alpha_i - \alpha_j|$ of vector $s$ encode the $\epsilon_r$-regularization and $\alpha$-gradient constraints.

The derivative of the cost function with respect to $g_i^c$ is then 
\begin{equation}
\frac{1}{2} \frac{\partial \cost}{\partial g_i^c} = (U_i^T g_i^c - I_i^c)U_i + R^T V_i (R^T g_i^c - h_i^c).
\end{equation}
Setting the derivative to zero and solving for $g_i^c$ yields the solution vector
\begin{equation} \label{eq:gsingle}
g_i^c = (U_i U_i^T + R^T V_i R)^{-1} (I_i^c U_i + R^T V_i h_i^c).
\end{equation}
The matrix $(U_i U_i^T + R^T V_i R)^{-1}$ is independent of $c$, which means that it only has to be computed once per pixel and can be reused for each color channel.

To solve the problem of slow propagation, we employ a multi-level approach. We begin by solving for the foreground image at a low resolution where the slow spatial propagation problem does not exist. It is sufficient to minimize the local cost function iteratively. Next, we solve the problem at a slightly larger scale by using the solution from the smaller scale as initialization. We repeat this process until the original size of the input image is reached.

\subsection{Implementation}

\begin{figure}
	\centering
	\includegraphics[width=8.3cm]{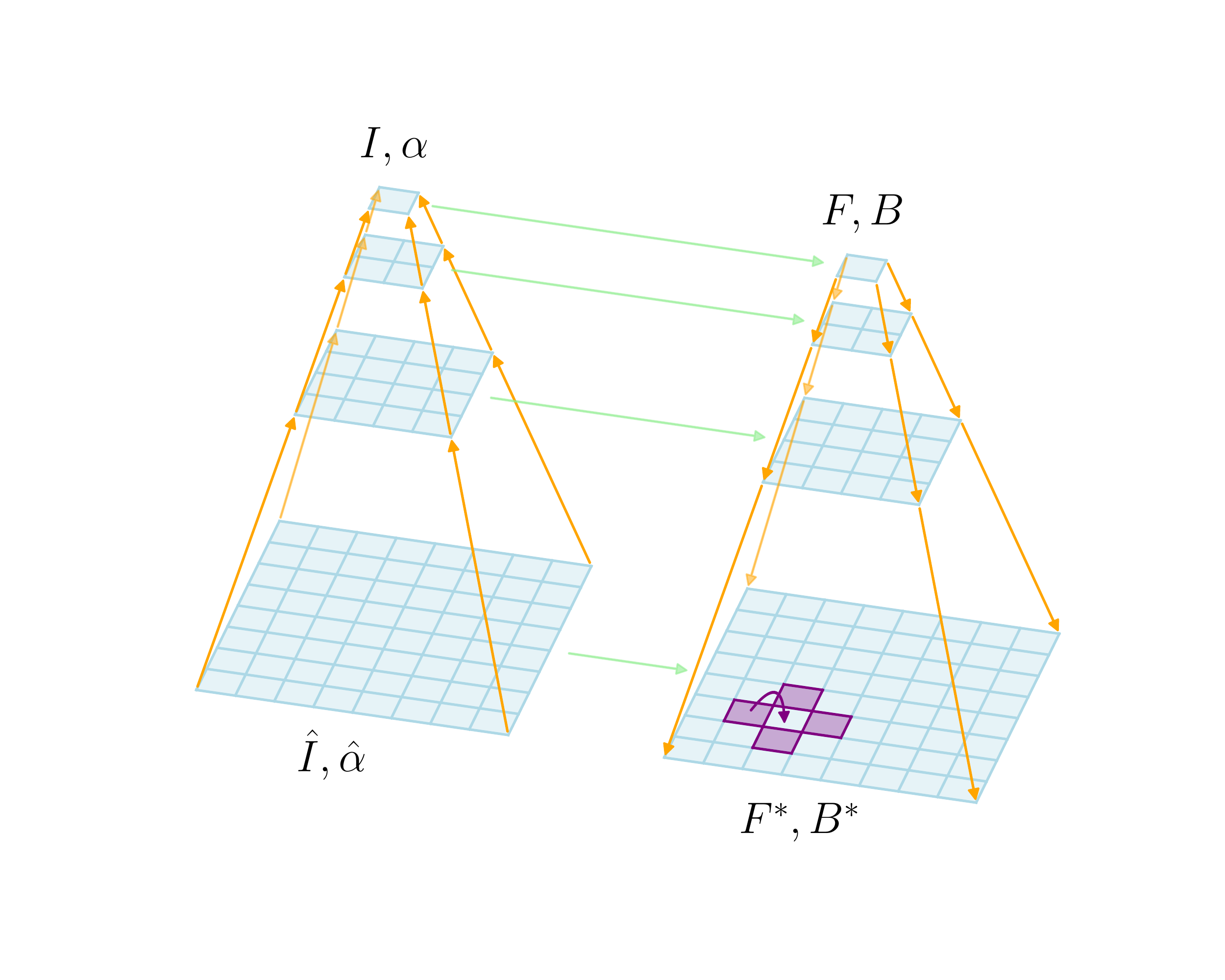}
	\caption{The full resolution input images $\hat{I}$ and $\hat{\alpha}$ are downsampled (left) and used to refine increasingly larger $F$ and $B$ images to obtain the solution $F^*$ and $B^*$ (right).}
	\label{fig:multilevel}
\end{figure}

The input to the multi-level color estimation procedure listed in Algorithm~\ref{alg:ml} is an RGB image $\hat{I} \in \left[0,1\right]^{\hat{w} \times \hat{h} \times 3}$ and an alpha matte $\hat{\alpha} \in \left[0,1\right]^{\hat{w} \times \hat{h}}$ of resolution $\hat{w} \times \hat{h}$ pixels.

At the smallest-level, foreground and background images $F$ and $B$ are initialized to a resolution of $1 \times 1$ pixels. For orientation, see the top of the right pyramid in Figure~\ref{fig:multilevel}.
The values at this point are not important since they will be updated later and converge quickly.

Next, a loop over the various image levels is started and the input image and input alpha matte, as well as the foreground and background images of the previous level, are resized to the current working size. The number of levels $n_l$ is chosen such that the image width and height grows at most by a factor of two between levels to ensure spatial propagation.

At each level and for each iteration, a linear system is constructed for each pixel coordinate $i = (x, y)$ from its neighbors $j = (x', y') \in~N_i$ (Figure~\ref{fig:multilevel}, purple, and Algorithm~\ref{alg:ml}, lines~10-24). Coordinates of neighbors which would exceed image bounds are clamped to the valid image region.

The linear system is then solved and applied simultaneously to update all color channels of the current pixel's foreground and background colors.

It is a good idea to run more iterations at lower resolutions because they are computationally cheaper and the colors are propagated further when the image is resized to a higher resolution. In practice, 2 iterations for medium to high resolutions and 10 iterations for low resolutions are usually sufficient to achieve visually pleasing results, where \textit{low} is chosen as $32 \times 32$ pixels.

\begin{algorithm}
	\caption{Multi-Level Color Estimation}
	\label{alg:ml}
	\begin{algorithmic}[1]
		
		\Function{\textproc{ML}}{$\hat{I}, \hat{\alpha}, \omega \gets 0.1, \epsilon_r \gets 5 \cdot 10^{-3}$}
		
		\State $F \gets 0_{1,1}, B \gets 0_{1,1}$ 
		
		\State $n_l \gets \ceil{\log_2({\max(\hat{w}, \hat{h})})}$
		\For{$l \gets 1$ \textbf{to} $n_l$}
		
		\State $\mathrlap{w}\hphantom{B} \gets \round({\hat{w}^{l/n_l}}), \mathrlap{h}\hphantom{B} \gets \round({\hat{h}^{l/n_l}})$
		\State $\mathrlap{I}\hphantom{B} \gets \resize(\hat{I}, w, h), \mathrlap{\alpha}\hphantom{B} \gets \resize(\hat{\alpha}, w, h)$
		\State $\mathrlap{F}\hphantom{B} \gets \resize(F, w, h), \mathrlap{B}\hphantom{B} \gets \resize(B, w, h)$
		
		\While{$F$ and $B$ not converged}
		
		\For{each pixel $(x, y) = i$ of $I$} 
		\State $U \gets [\alpha_i, 1 - \alpha_i]^T$
		\State $A \gets U U^T$
		\State $b \gets U I_i^T$
		\For{each neighbor$(x', y') \in N_i$}
		\State $x' \gets \max(0, \min(w - 1, x'))$ 
		\State $\hphantom{x'}\mathllap{y'} \gets \max(0, \min(\hphantom{w}\mathllap{h} - 1, \hphantom{x'}\mathllap{y'}))$ 
		\State $j \gets (x', y')$
		
		\State $\Delta_\alpha \gets \epsilon_r + \omega |\alpha_i - \alpha_j|$ 
		
		\State $A_{1,1} \gets A_{1, 1} + \Delta_\alpha$
		\State $A_{2,2} \gets A_{2, 2} + \Delta_\alpha$
		
		\For{$c \gets 1$~\textbf{to}~$3$} 
		\State $b_{1,c} \gets b_{1,c} + \Delta_\alpha F_j^c$
		\State $b_{2,c} \gets b_{2,c} + \Delta_\alpha B_j^c$
		\EndFor
		
		\EndFor
		
		\State $g \gets A^{-1} b$ 
		
		\For{$c \gets 1$ \textbf{to} $3$} 
		\State $F_i^c \gets \max(0, \min(1, g_{1,c})$ 
		\State $B_i^c \gets \max(0, \min(1, g_{2,c})$
		\EndFor
		
		\EndFor
		\EndWhile
		\EndFor
		\State \Return{$F, B$}
		\EndFunction
	\end{algorithmic}
\end{algorithm}

\section{Experiments}

\begin{figure*}
	\centering
	\includegraphics[clip, width=\textwidth]{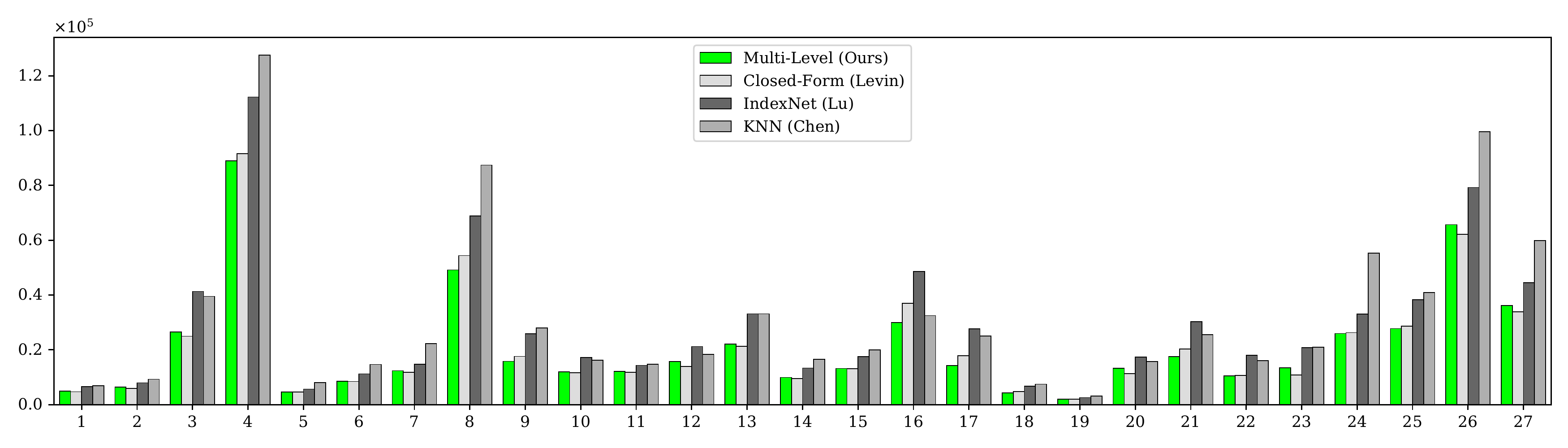} 
	\caption{The sum of absolute differences (SAD) between the ground truth foreground  $F^{\text{gt}}$ and the estimated foreground $F^{\text{est}}$ computed from the ground truth alpha matte $\alpha^{gt}$ weighted by $\alpha^{gt}$ over the translucent region (Equation~\ref{eq:sad}) for all 27 images of the dataset by \cite{rhemann2009perceptually} for four foreground estimation methods.}
	\label{fig:sad}
\end{figure*}
\begin{figure*}
	\centering
	\includegraphics[clip, width=\textwidth]{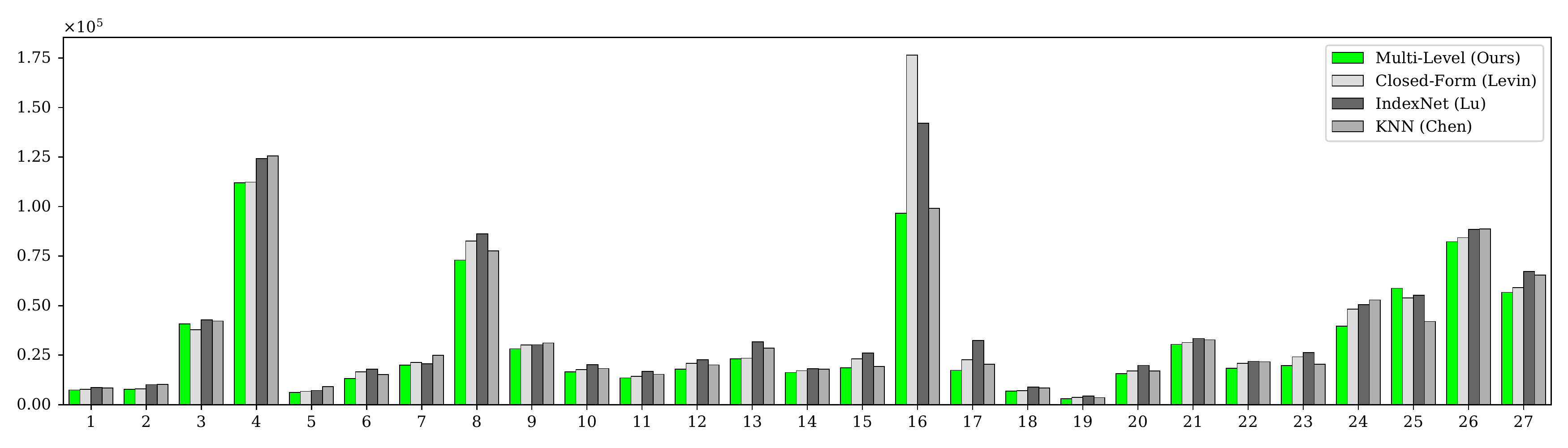} 
	\caption{The SAD error for the foreground images computed from the alpha matte estimated with KNN alpha matting.}
	\label{fig:sad-knn}
\end{figure*}

\subsection{Other Methods}

We compare the quality of the estimated foreground of our method with that of the author's implementations of closed-form foreground estimation \cite{levin2007closed} and KNN foreground estimation \cite{chen2013knn} as well as the ground truth on the dataset \cite{rhemann2009perceptually}.

Other methods exist which estimate both alpha and foreground given an input image and a trimap, for example using neural networks \cite{hou2019context, tang2019learning}. However, those methods are not easily comparable since on the one hand, an alpha matte is a more precise input than a trimap, making this a harder problem, but on the other hand, those methods could also trade off error in the alpha matte against error in the foreground estimation. Therefore, we do not include such methods in our evaluation.

Nevertheless, comparing against a neural network-based foreground estimation method is conceptually interesting, which is why we modify the IndexNet \cite{lu2019indices} alpha matting network to predict a foreground estimate instead. We retrained the network on the ground truth dataset by \cite{xu2017deep} using a compositional loss on the unknown image region
\begin{equation}
loss(F^{\text{est}})=\sum_{0 < \alpha_i^{\text{gt}} < 1} \left|\alpha_i^{\text{gt}} F_i^{\text{est}} + (1 - \alpha_i^{\text{gt}}) B_i^{\text{gt}} - I_i^{\text{gt}}\right|_1. 
\end{equation}

We otherwise adapt the same training procedure as described by \cite{lu2019indices}.

To compare the runtime of our multi-level approach with the other methods, we implement KNN foreground estimation and closed-form foreground estimation using Python with vectorized NumPy and SciPy routines.

\subsection{Dataset}

\begin{figure}
	\centering
	\begin{tabular}{cccccc} 
		\includegraphics[width=2cm, align=c]{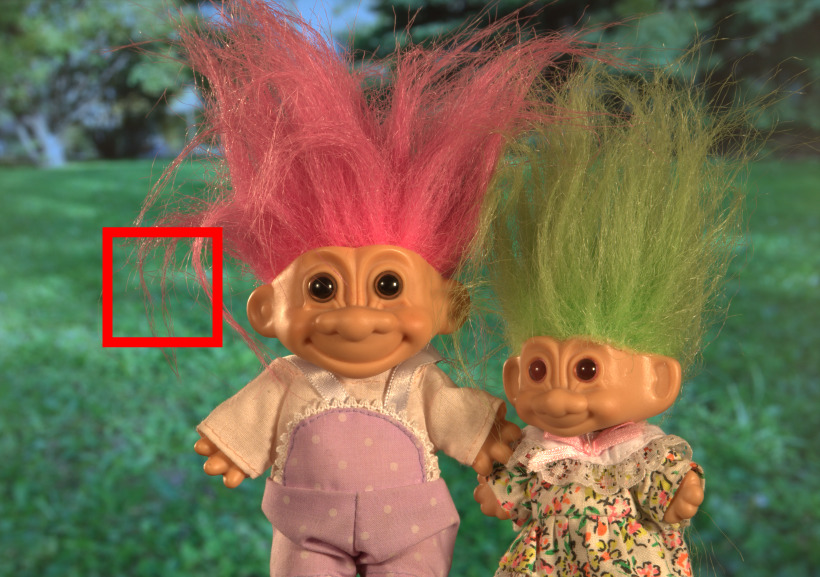} &
		\includegraphics[width=2cm, align=c]{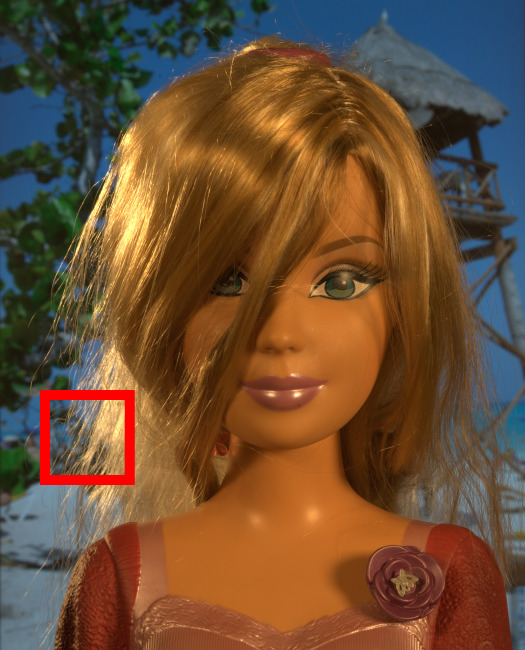} &
		\includegraphics[width=2cm, align=c]{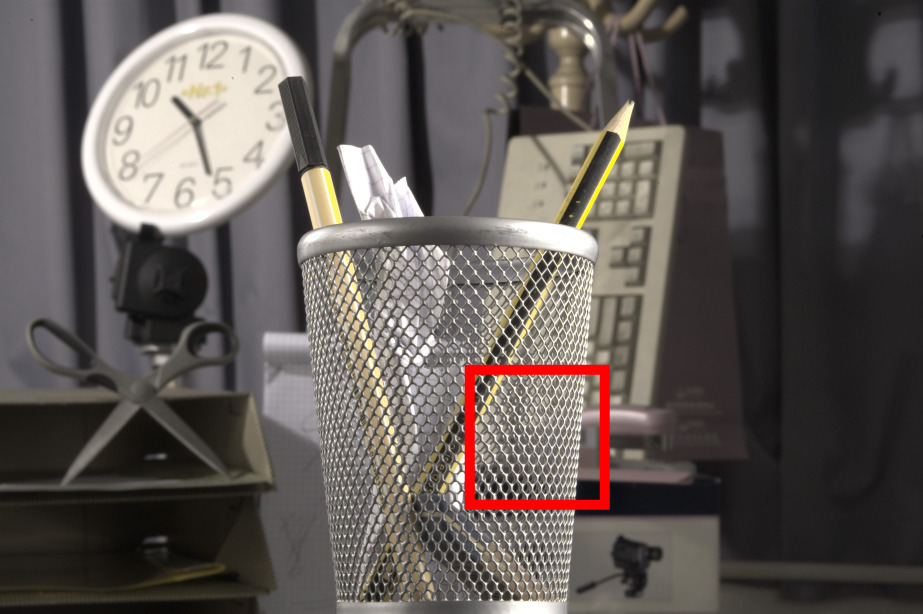} 
	\end{tabular}
	\caption{Images from the dataset \cite{rhemann2009perceptually}. The red boxes indicate challenging image regions which we will focus on later.}
	\label{fig:dataset}%
\end{figure}

We use the high resolution input images, alpha mattes and foreground images of the dataset by Rhemann et al. \cite{rhemann2009perceptually} to evaluate the quality of the foreground estimation methods. A selected sample of images is displayed in Figure~\ref{fig:dataset}.

It should be noted that, due to noise in the data, the ground truth foreground color does not exactly match the color of the corresponding input image, even in regions where the alpha matte is equal to one. 
Thus, having exactly zero error is not possible for the given data.

The alpha matting dataset by \cite{rhemann2009perceptually} contains ground truth images $I_{\text{lrgb}}$ and foreground images $F_{\text{lrgb}}$ in linear RGB color space without white point, as well as images $I^{'W}_{\text{srgb}}$ in sRGB color space with white point adjustment.
Although it would be physically more accurate to apply methods in linear RGB space, it is more common to use the sRGB color space instead. Therefore, we transform the linear RGB ground truth foreground images without white point to sRGB space with white point correction.

The white point parameters are unknown. For this reason, we employ an optimization approach to obtain a matrix $M$ to transform from $I_{\text{lrgb}}$ into $I^{'W}_{\text{lrgb}}$, denoted by $n$-by-3 matrices $V$ and $W$ of stacked linear RGB color row vectors respectively.
To get $I^{'W}_{\text{lrgb}}$ from $I^{'W}_{\text{srgb}}$, we apply the inverse gamma correction function 
\begin{equation}
f^{-1}(s)={\begin{cases}\frac{s}{12.92} & s \leq 0.04045\\
	\left({\tfrac {s + 0.055}{1.055}}\right)^\gamma & {\text{otherwise}}\end{cases}}
\end{equation}
element-wise to each entry and color value.
We minimize the error function
\begin{equation}
\err_{\text{lrgb}}(M) = \sum \left[ (MV^T - W^T) \odot (MV^T - W^T)\right],
\end{equation}
arriving at the 3-by-3 white point correction matrix $M$
\begin{equation}
M = (W^T V)(V^T V)^{-1}.
\end{equation}
We can thus obtain $F^W_{\text{lrgb}}$ by multiplying $M$ with each 3-by-1 color vector of $F_{\text{lrgb}}$ and finally applying the gamma correction function
\begin{equation}
f(l)={\begin{cases}12.92~l & l \leq 0.0031308\\
	1.055~l^{1/\gamma} - 0.055 & {\text{otherwise}}\end{cases}}
\end{equation}
to transform from $F^W_{\text{lrgb}}$ to $F^W_{\text{srgb}}$, corresponding to the ground truth $F^{\text{gt}}$. We compare the output of the various methods against $F^{\text{gt}}$. Likewise, we transform $I_{\text{lrgb}}$ to $I^W_{\text{srgb}}$, which corresponds to the input image $I$.
It should be noted that the image $I^{'W}_{\text{srgb}}$ can not be used in place of $I^W_{\text{srgb}}$ because, although the images are often similar, notable differences exist. For example, the background in image 12 of the dataset is different.
The gamma parameter was chosen as $\gamma = 2$ to minimize $\text{err}_{\text{lrgb}}$ over all images.

We deliberately chose not to evaluate the methods on the Composition 1k dataset because the characteristic artifacts suggest that the foreground images were obtained using a variation of closed-form color estimation, which would give an unfair advantage to methods which are conceptually similar, since they are likely to exhibit the same artifacts.
On the other hand, the foreground images in the dataset \cite{rhemann2009perceptually} were computed from multiple photos with varying backgrounds, which should not bias the evaluation towards specific methods.

\subsection{Error Measures}

We report the sum of absolute differences (SAD) between the ground truth foreground  $F^{\text{gt}}$ and the estimated foreground $F^{\text{est}}$ weighted by the ground truth alpha matte $\alpha^{\text{gt}}$ over the translucent region  
\begin{equation}
d_{\text{SAD}}(F^{\text{est}}, F^{\text{gt}})=\sum_{0 < \alpha_i^{\text{gt}} < 1} \alpha_i^{\text{gt}} \left|F_i^{\text{est}} - F_i^{\text{gt}}\right|_1.
\label{eq:sad}
\end{equation}
We also report the mean squared error (MSE) which we weight similarly:
\begin{equation}
d_{\text{MSE}}(F^{\text{est}}, F^{\text{gt}})=\sum_{0 < \alpha_i^{\text{gt}} < 1} \alpha_i^{\text{gt}} \left|F_i^{\text{est}} - F_i^{\text{gt}}\right|^2_2. 
\end{equation}
Furthermore, Rhemann et al. \cite{rhemann2009perceptually} show that the gradient error of the alpha matte correlates with the perceptual quality. We adapt it to compute the error of the gradient of the foreground (GRAD) as
\begin{equation}
d_{\text{GRAD}}(F^{\text{est}}, F^{\text{gt}})=\sum_{0 < \alpha_i^{\text{gt}} < 1} \alpha_i^{\text{gt}} \left|\nabla F_i^{\text{est}} - \nabla F_i^{\text{gt}}\right|^2_2,
\end{equation}
where $\nabla F$ denotes the gradient image $F$, which is calculated by first-order Gaussian derivative filters with standard deviation $\sigma = 1.4$.

\subsection{Qualitative Results}

\begin{figure}[ht]
    \vspace{1cm}
    \centering
    \begin{tabular}{rccccccc}
        Input Image &
        \includegraphics[width=\imgwidth, align=c]{figures/zoom_new/image/ifm/GT04.jpg} &
        \includegraphics[width=\imgwidth, align=c]{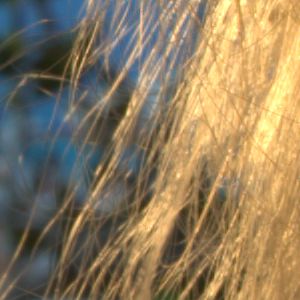} &
        \includegraphics[width=\imgwidth, align=c]{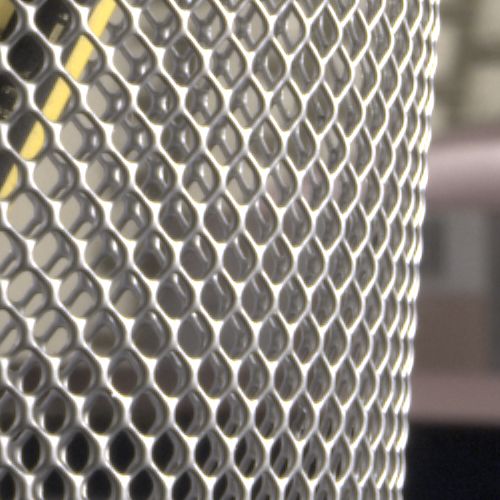} \\
        \vspace{\vpad}\\
        Input Alpha &
        \includegraphics[width=\imgwidth, align=c]{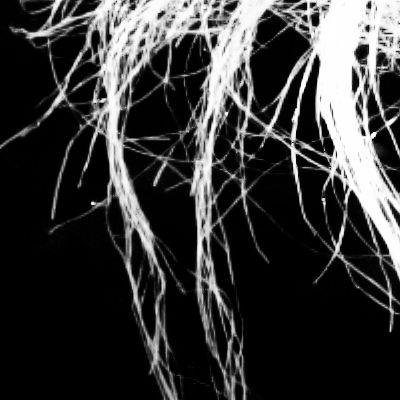} &
        \includegraphics[width=\imgwidth, align=c]{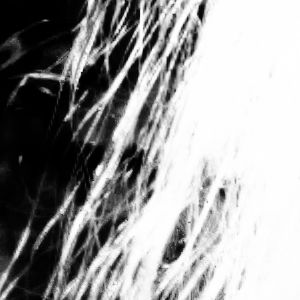} &
        \includegraphics[width=\imgwidth, align=c]{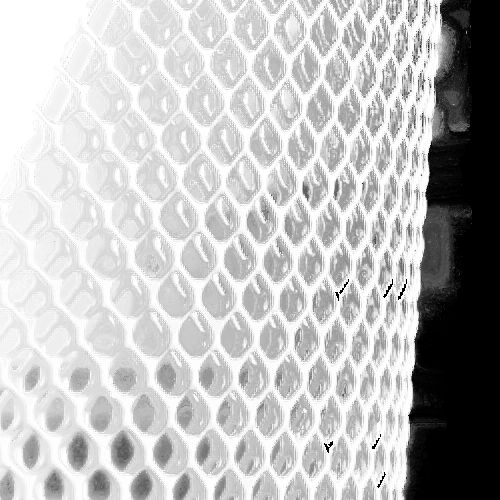} \\
        \vspace{\vpad}\\
        CF &
        \includegraphics[width=\imgwidth, align=c]{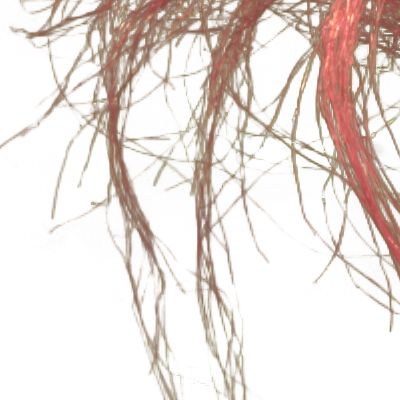} &
        \includegraphics[width=\imgwidth, align=c]{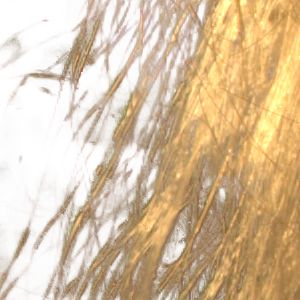} &
        \includegraphics[width=\imgwidth, align=c]{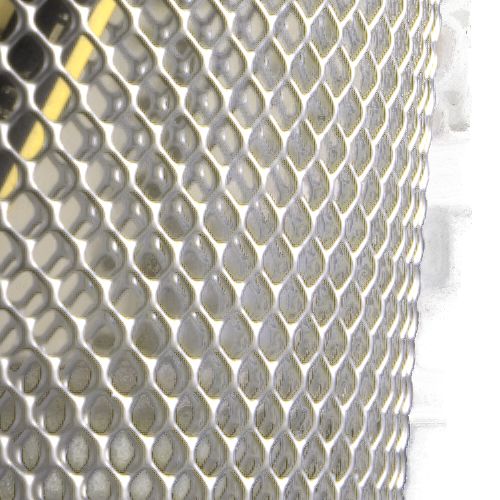} \\
        \vspace{\vpad}\\
         IndexNet &
        \includegraphics[width=\imgwidth, align=c]{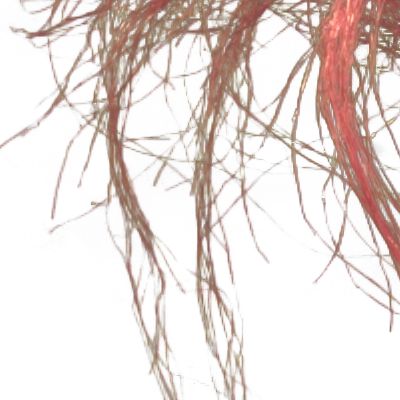} &
        \includegraphics[width=\imgwidth, align=c]{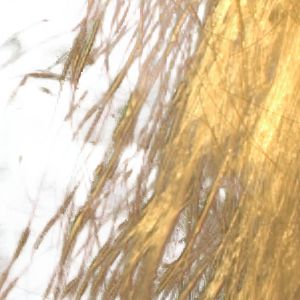} &
        \includegraphics[width=\imgwidth, align=c]{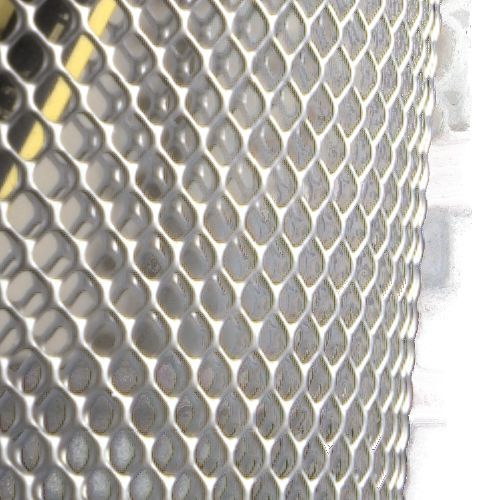} \\
        \vspace{\vpad}\\
        KNN &
        \includegraphics[width=\imgwidth, align=c]{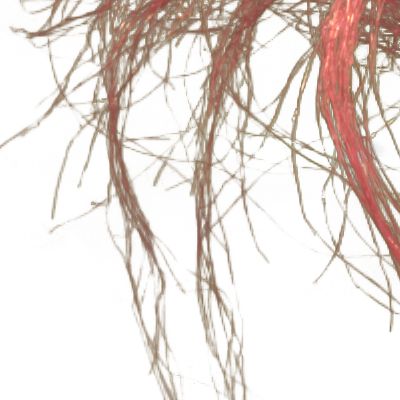} &
        \includegraphics[width=\imgwidth, align=c]{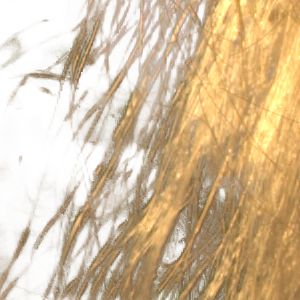} &
        \includegraphics[width=\imgwidth, align=c]{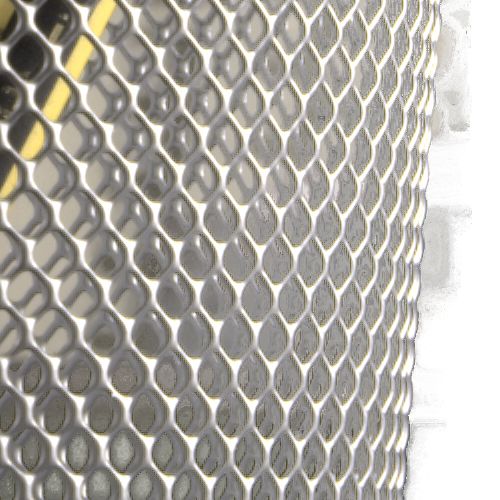} \\
        \vspace{\vpad}\\
        ML (Ours) &
        \includegraphics[width=\imgwidth, align=c]{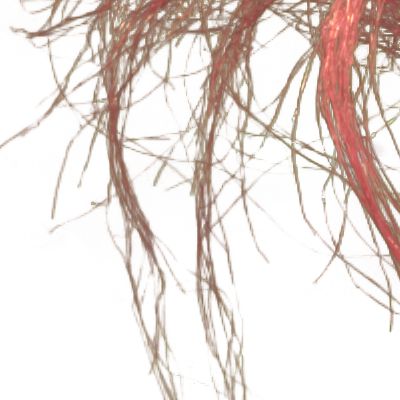} &
        \includegraphics[width=\imgwidth, align=c]{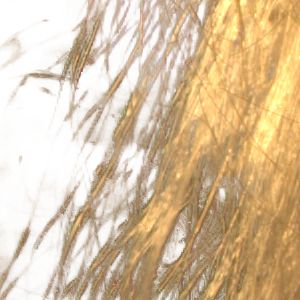} &
        \includegraphics[width=\imgwidth, align=c]{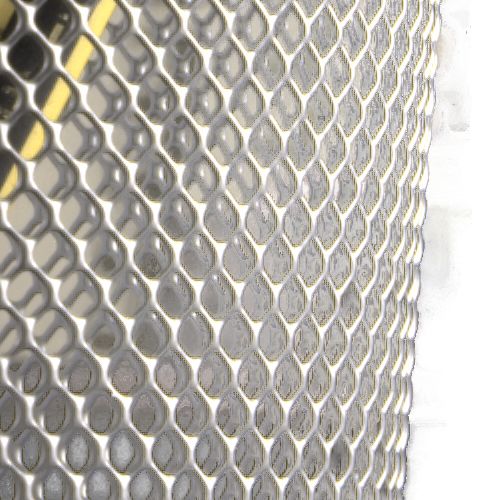} 
    \end{tabular}
    \caption{Results for each foreground estimation method on the alpha matte estimated with KNN matting. Best viewed on a digital display.}
    \label{fig:knncomposite}%
\end{figure}

\begin{figure}[ht]
    \vspace{1cm}
    \centering
    \begin{tabular}{rccccccc}
        Image  &
        \includegraphics[width=\imgwidth, align=c]{figures/zoom_new/image/ifm/GT04.jpg} &
        \includegraphics[width=\imgwidth, align=c]{figures/zoom_new/image/ifm/GT08.jpg} &
        \includegraphics[width=\imgwidth, align=c]{figures/zoom_new/image/ifm/GT25.jpg} \\
        \vspace{\vpad}\\
        Alpha&
        \includegraphics[width=\imgwidth, align=c]{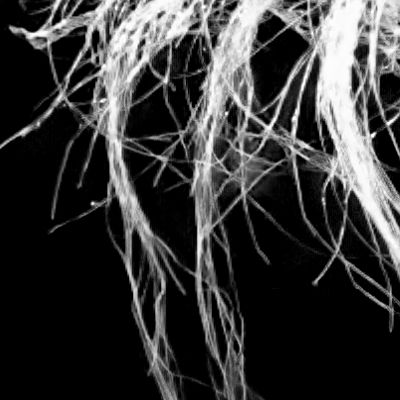} &
        \includegraphics[width=\imgwidth, align=c]{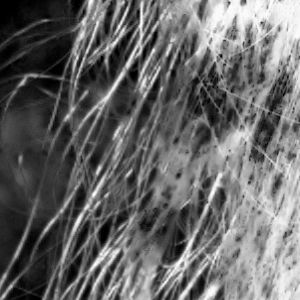} &
        \includegraphics[width=\imgwidth, align=c]{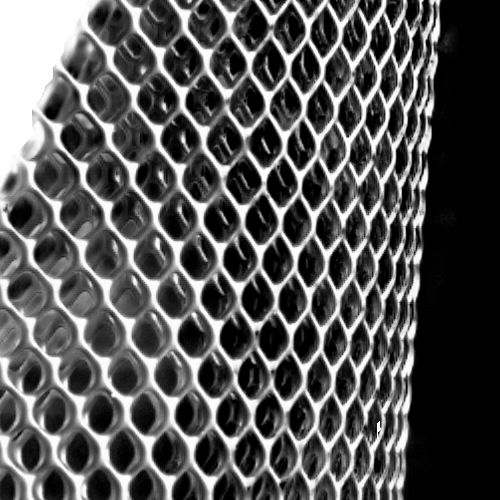} \\
        \vspace{\vpad}\\
         CF&
        \includegraphics[width=\imgwidth, align=c]{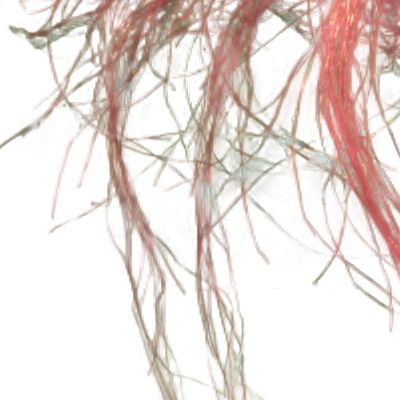} &
        \includegraphics[width=\imgwidth, align=c]{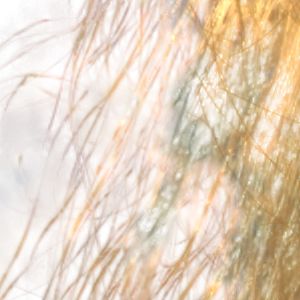} &
        \includegraphics[width=\imgwidth, align=c]{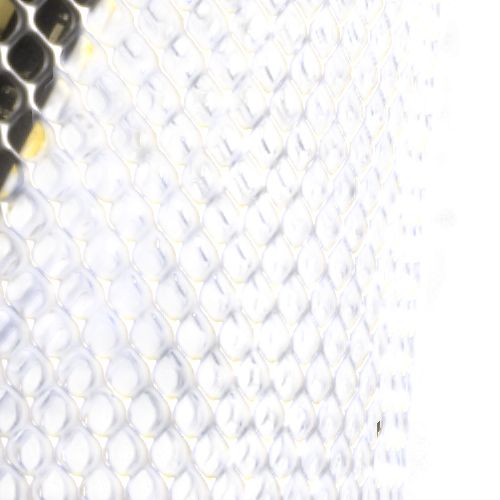} \\
        \vspace{\vpad}\\
        IndexNet &
        \includegraphics[width=\imgwidth, align=c]{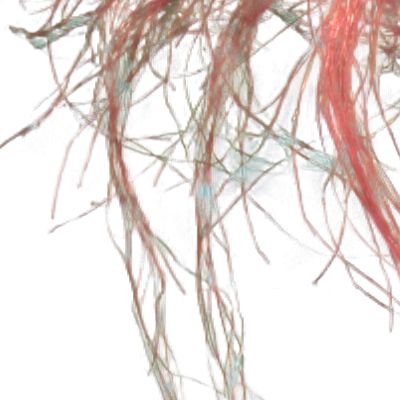} &
        \includegraphics[width=\imgwidth, align=c]{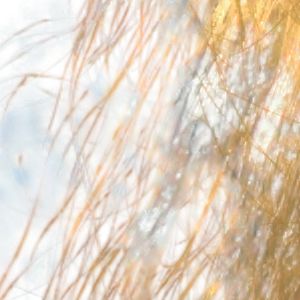} &
        \includegraphics[width=\imgwidth, align=c]{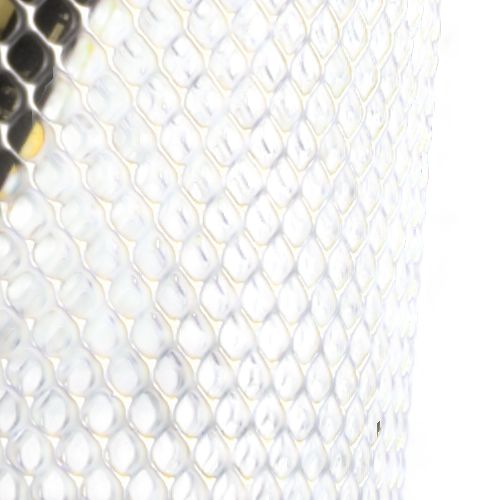} \\
        \vspace{\vpad}\\
        KNN &
        \includegraphics[width=\imgwidth, align=c]{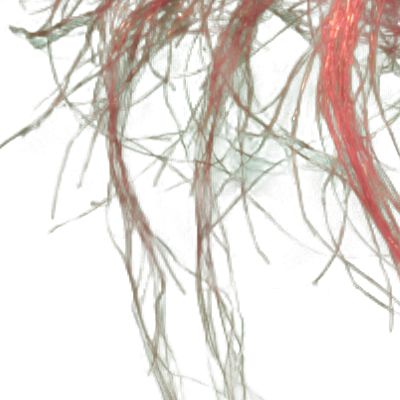} &
        \includegraphics[width=\imgwidth, align=c]{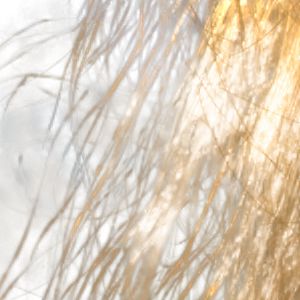} &
        \includegraphics[width=\imgwidth, align=c]{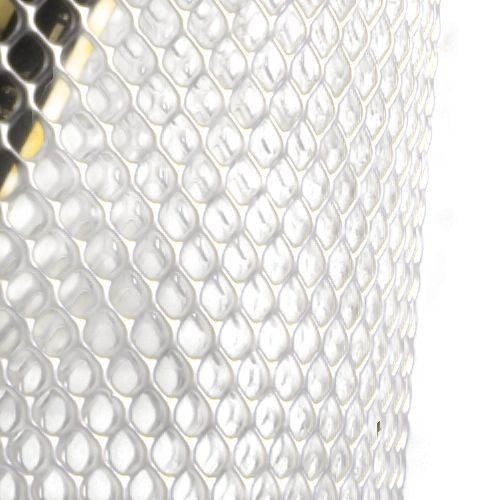} \\
        \vspace{\vpad}\\
        ML &
        \includegraphics[width=\imgwidth, align=c]{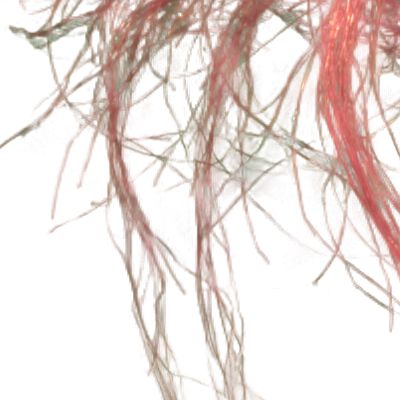} &
        \includegraphics[width=\imgwidth, align=c]{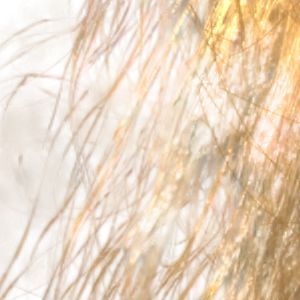} &
        \includegraphics[width=\imgwidth, align=c]{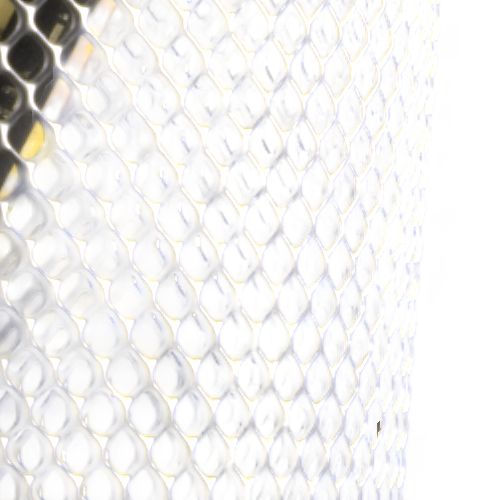} 
    \end{tabular}
    \caption{Results for each foreground estimation method on the alpha matte estimated with IndexNet.}
    \label{fig:idxcomposite}%
\end{figure}

\begin{figure}[ht]
    \vspace{1cm}
    \centering
    \begin{tabular}{rccccccc}
            
        Input &
        \includegraphics[width=\imgwidth, align=c]{figures/zoom_new/image/ifm/GT04.jpg} &
        \includegraphics[width=\imgwidth, align=c]{figures/zoom_new/image/ifm/GT08.jpg} &
        \includegraphics[width=\imgwidth, align=c]{figures/zoom_new/image/ifm/GT25.jpg} \\
        \vspace{\vpad}\\
        Alpha &
        \includegraphics[width=\imgwidth, align=c]{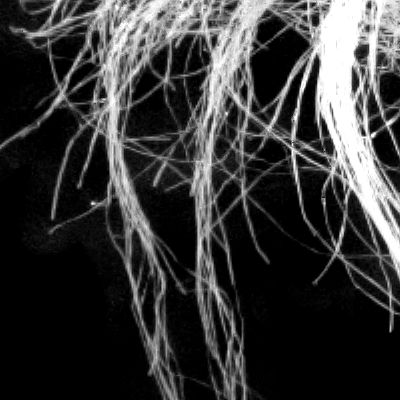} &
        \includegraphics[width=\imgwidth, align=c]{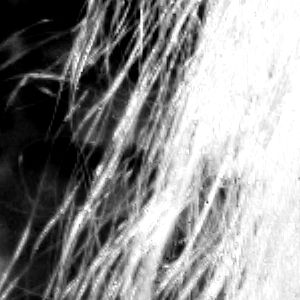} &
        \includegraphics[width=\imgwidth, align=c]{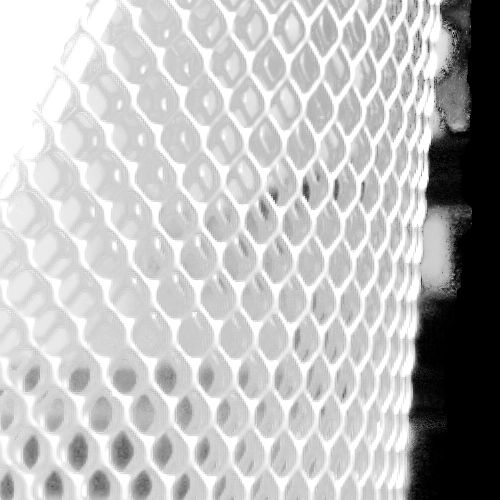} \\
        \vspace{\vpad}\\
        CF &
        \includegraphics[width=\imgwidth, align=c]{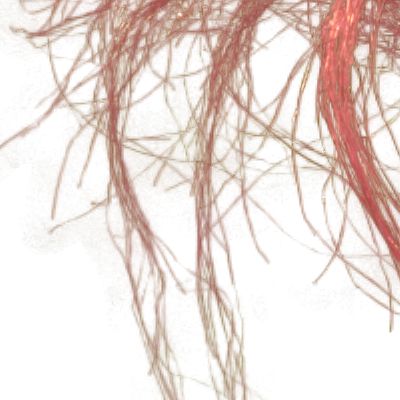} &
        \includegraphics[width=\imgwidth, align=c]{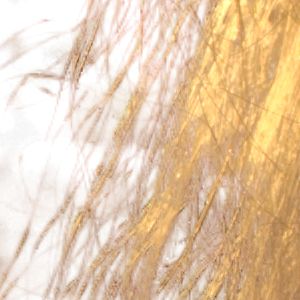} &
        \includegraphics[width=\imgwidth, align=c]{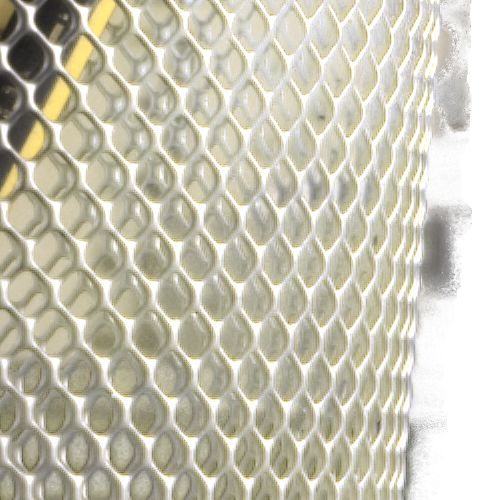} \\
        \vspace{\vpad}\\
        IndexNet &
        \includegraphics[width=\imgwidth, align=c]{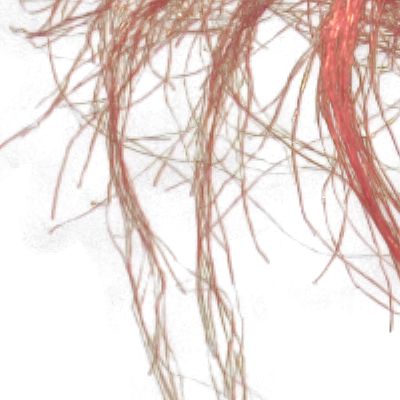} &
        \includegraphics[width=\imgwidth, align=c]{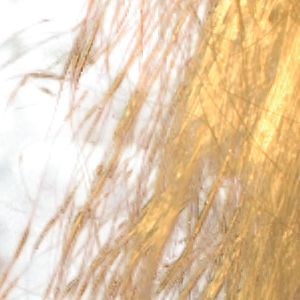} &
        \includegraphics[width=\imgwidth, align=c]{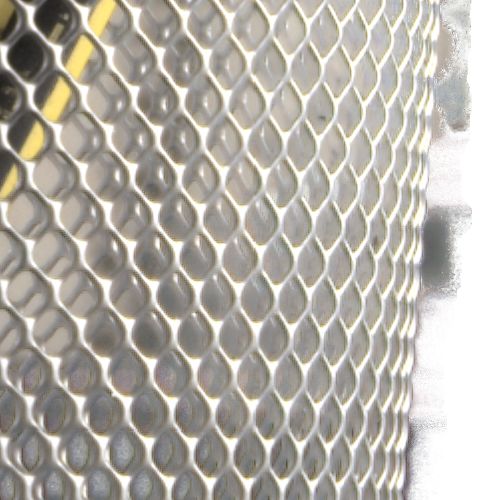} \\
        \vspace{\vpad}\\
        KNN &
        \includegraphics[width=\imgwidth, align=c]{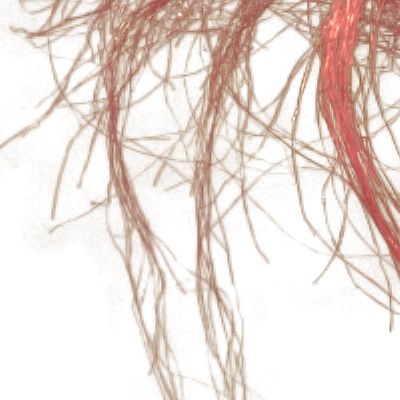} &
        \includegraphics[width=\imgwidth, align=c]{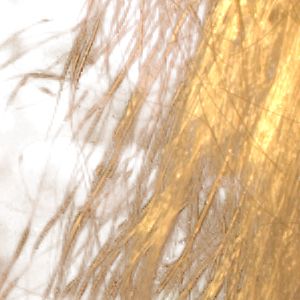} &
        \includegraphics[width=\imgwidth, align=c]{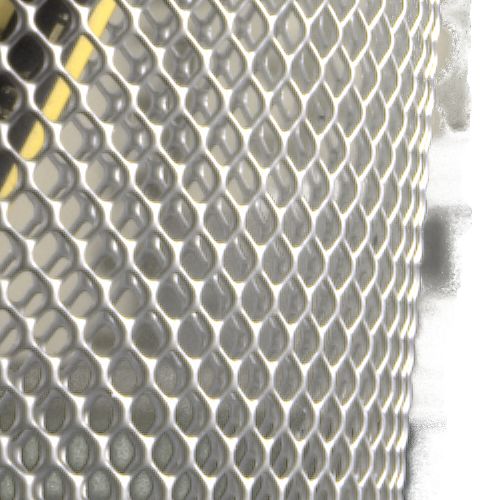} \\
        \vspace{\vpad}\\
        ML (Ours) &
        \includegraphics[width=\imgwidth, align=c]{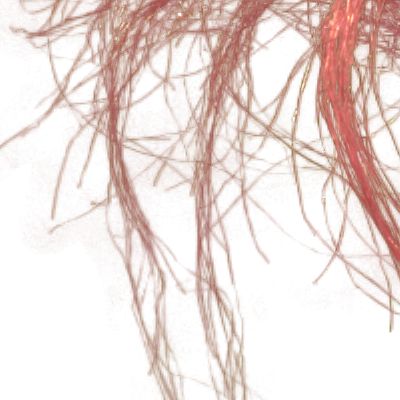} &
        \includegraphics[width=\imgwidth, align=c]{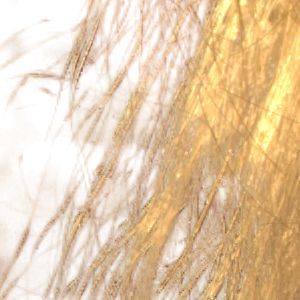} &
        \includegraphics[width=\imgwidth, align=c]{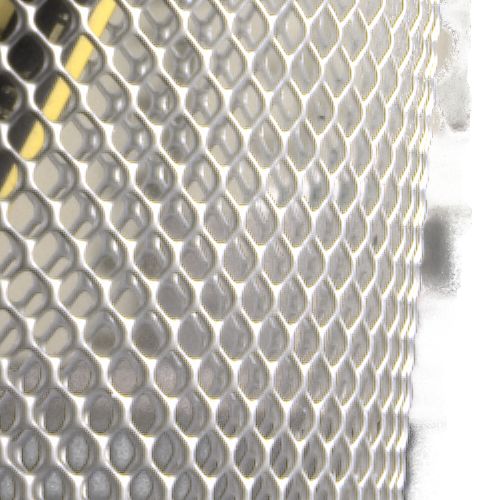}
    \end{tabular}
    \caption{Results for each foreground estimation method on the alpha matte estimated with information flow matting.}
    \label{fig:ifmcomposite}%
\end{figure}

We are mainly interested in the quality of the estimated foreground images where the ground truth alpha matte is not available. For this reason, we compute several alpha mattes using the respective author's implementation of KNN matting \cite{chen2013knn}, IndexNet matting \cite{lu2019indices} and information-flow matting \cite{ifm} (Figure~\ref{fig:knncomposite},~\ref{fig:idxcomposite},~\ref{fig:ifmcomposite}, second row). The alpha values estimated by KNN alpha matting are often close to zero or one. Information-flow matting produces smoother values. IndexNet matting seems to be struggling with the mesh structure, which consists of quickly alternating bright and dark colors due to reflection.

\begin{figure*}[ht]
	\centering
	\begin{tabular}{c|cccc|c}
		\includegraphics[width=\imgwidth]{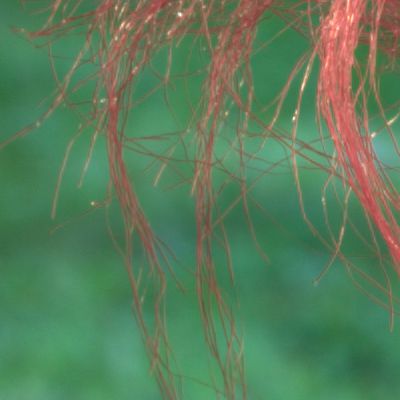} 
	    \hspace{0.2cm}&\hspace{0.2cm}
		\includegraphics[width=\imgwidth]{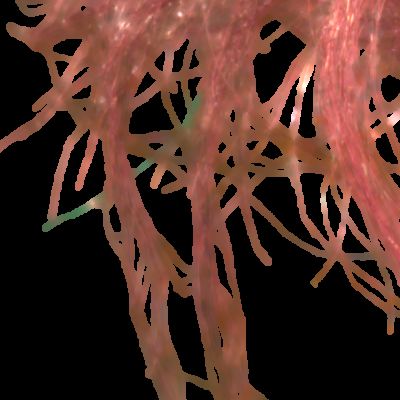} &
		\includegraphics[width=\imgwidth]{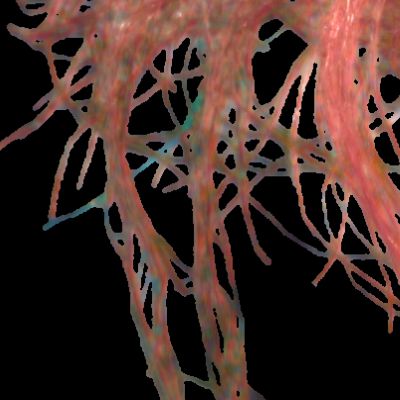} &
		\includegraphics[width=\imgwidth]{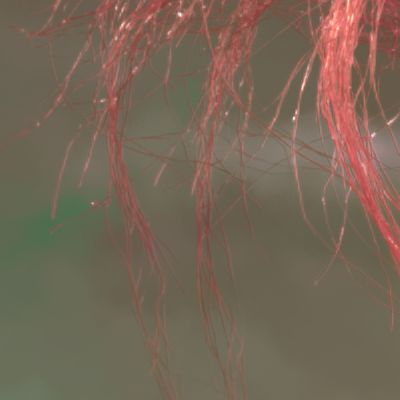} &
		\includegraphics[width=\imgwidth]{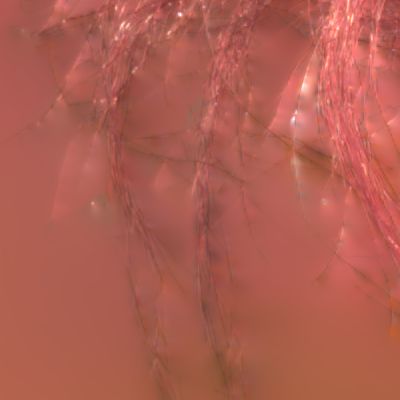} 
		\hspace{0.2cm}&\hspace{0.2cm}
		\includegraphics[width=\imgwidth]{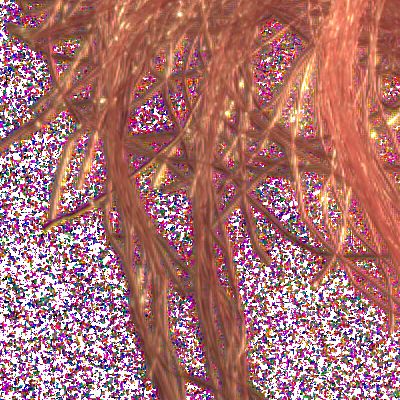}\\
		Input  & CF & IndexNet & KNN & ML & GT
		
	\end{tabular}
	\caption{Input image, estimated foreground based on closed-form, IndexNet, KNN and multi-level foreground estimation followed by the ground truth.}
	\label{fig:foreground}%
\end{figure*}

First, we discuss the raw estimated foreground colors without composing them onto a background (Figure~\ref{fig:foreground}). It can be seen that both closed-form foreground estimation and the retrained IndexNet do not propagate the foreground color far into the background region. While this is not an issue when compositing the foreground with the ground truth alpha matte, it could be an issue for incorrectly estimated alpha mattes. KNN foreground estimation propagates the colors further, resulting in a background which is slightly tinted with the foreground colors. Lastly, our multi-level method strongly propagates colors, producing a foreground estimate which is suitable even for inaccurate alpha estimates.

To evaluate the qualitative results, we compose the estimated foreground images onto a white background to make it easier to see if any traces of the background color are showing through and have not been removed satisfactorily.

In the case where the KNN alpha matte is used as input, the estimated foreground colors are usually too dark due to the almost-binary nature of the alpha matte. This can be observed across all tested foreground estimation methods (Figure~\ref{fig:knncomposite}).

For the IndexNet alpha matte, the green and blue background color still shines through due to artifacts in the alpha matte. This effect is greatly diminished for our method due to the strong propagation of foreground colors into background regions (Figure~\ref{fig:idxcomposite}, last row).

Information-flow alpha matting slightly overestimates the alpha matte for the wire mesh image (Figure~\ref{fig:ifmcomposite}, third column), resulting in a green-colored mesh for closed-form foreground estimation as well as dark blotches for the other methods. Otherwise, all methods produce acceptable results.

\begin{table}
	\centering
	\caption{Sum of absolute differences, mean squared error and gradient error \cite{rhemann2009perceptually} for different alpha mattes averaged over the dataset by \cite{rhemann2009perceptually}.}
	\begin{tabular}{llrrr}
		\toprule
		Alpha & Foreground  & SAD              & MSE             & GRAD             \\
		      &             & \tiny{$10^{-3}$} & \tiny{$10^{3}$} & \tiny{$10^{-3}$} \\
		\midrule
		\multirow{4}{1cm}{$\alpha_{gt}$}
		 & Multi-Level (Ours) & \textbf{20.9} & 1.44 & 8.89\\
		 & Closed-Form (Levin) & 21.1 & \textbf{1.34} & \textbf{8.13}\\
		 & IndexNet (Lu) & 28.8 & 2.33 & 11.1\\
		 & KNN (Chen) & 32.0 & 3.25 & 16.1\\
		\midrule
		\multirow{4}{1cm}{$\alpha_{KNN}$}
		 & Multi-Level (Ours) & \textbf{31.8} & \textbf{2.5} & \textbf{11.5}\\
		 & Closed-Form (Levin) & 36.6 & 3.51 & 14.2\\
		 & IndexNet (Lu) & 38.3 & 3.9 & 14.5\\
		 & KNN (Chen) & 34.6 & 3.22 & 13.0\\
		\midrule
		\multirow{4}{1cm}{$\alpha_{IDX}$}
		 & Multi-Level (Ours) & 47.9 & 5.66 & \textbf{15.8}\\
		 & Closed-Form (Levin) & 59.0 & 8.03 & 21.5\\
		 & IndexNet (Lu) & 62.6 & 8.65 & 21.4\\
		 & KNN (Chen) & \textbf{37.1} & \textbf{3.81} & 16.9\\
		\midrule
		\multirow{4}{1cm}{$\alpha_{IFM}$}
		 & Multi-Level (Ours) & \textbf{31.6} & \textbf{2.44} & \textbf{11.4}\\
		 & Closed-Form (Levin) & 37.7 & 3.98 & 15.3\\
		 & IndexNet (Lu) & 36.4 & 3.93 & 15.7\\
		 & KNN (Chen) & 33.7 & 2.97 & 13.6\\
		\bottomrule
	\end{tabular}
    \label{tab:errors}%
\end{table}

\subsection{Quantitative Results}

Figure~\ref{fig:sad} and Figure~\ref{fig:sad-knn} visualize the sum of absolute differences (Equation~\ref{eq:sad}) of the estimated foreground for each method applied to the ground truth alpha matte and KNN alpha matte respectively. The plots show that our method produces small errors not only when being applied to the ground truth, but also in the more realistic case when the alpha matte needs to be estimated.

Table~\ref{tab:errors} shows the SAD, MSE and Gradient error measures averaged over the dataset by Rhemann et al. \cite{rhemann2009perceptually}. Our multi-level method performs best with respect to SAD and gradient error for three of the four input alpha mattes. We point out that SAD is more perceptually relevant compared to MSE for image similarity \cite{sinha2011perceptually} and image restoration \cite{zhao2016loss}. The gradient error has been shown to be superior to both measures in the case of alpha matting \cite{rhemann2009perceptually}.

\subsection{Influence of Regularization}

\begin{figure}
	\centering
	\includegraphics[width=0.45\textwidth]{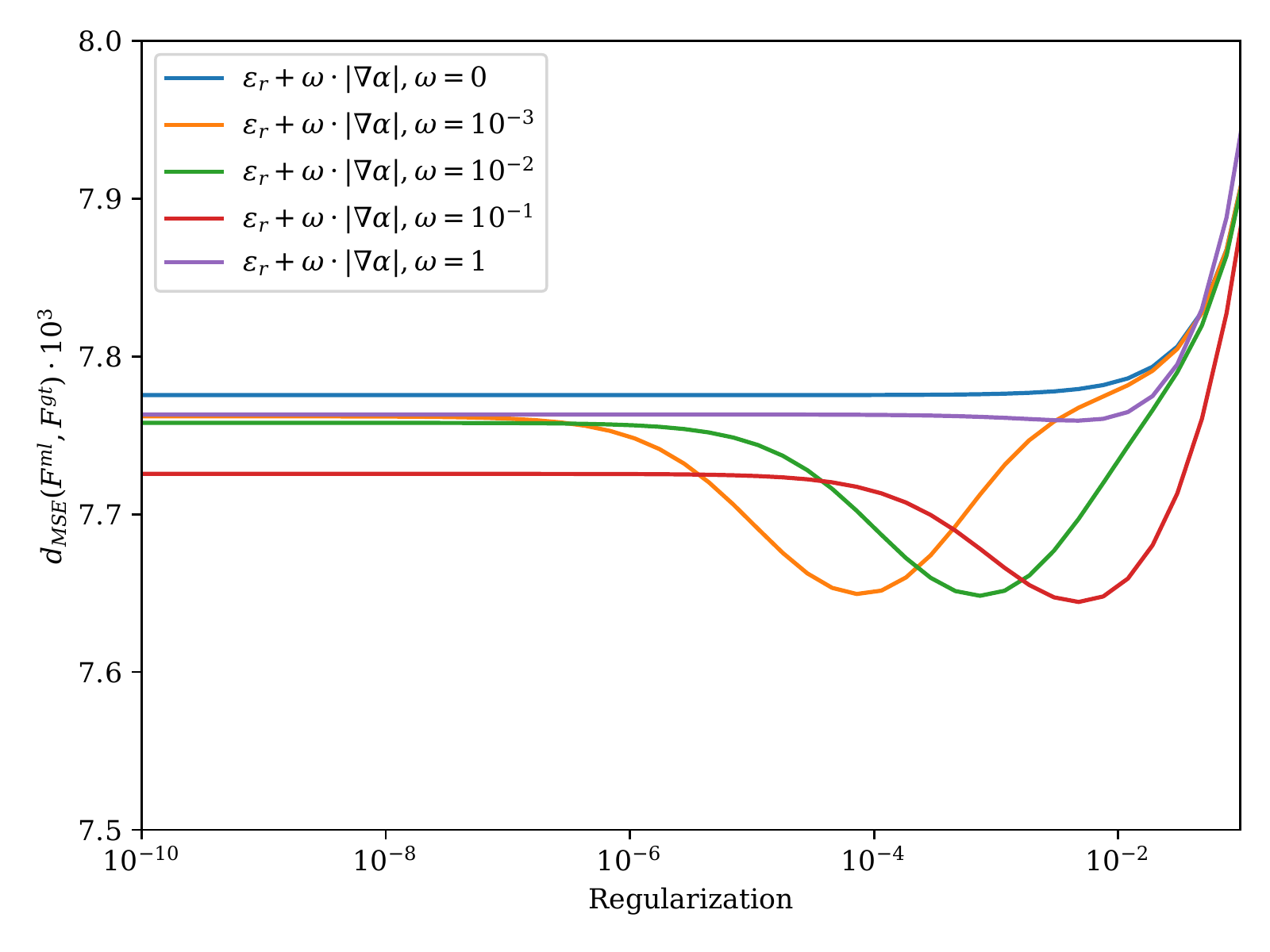} 
	\caption{Mean squared error between the ground truth foreground and foreground estimated with the multi-level method on ground truth alpha $\alpha_{gt}$.}
	\label{fig:regularization_ml_plot}
\end{figure}

We evaluate the influence of regularization on the error of the estimated foreground color (Figure~\ref{fig:regularization_ml_plot}) and make two key observations.

Firstly, the alpha gradient term by itself does not contribute much to the overall mean squared error, since the difference between weighting it with either $\omega = 0$ or $\omega = 1$ is small.

Secondly, a pronounced minimum exists with respect to the regularization factor when the alpha gradient term has a small but non-zero contribution.

Based on those observations, we choose a regularization factor of $\epsilon_r = 5 \cdot 10^{-3}$ and weight the alpha gradient term by $\omega = 0.1$ for all experiments.

\subsection{Runtime and Memory Usage}

\begin{table}
	\centering
	\caption{Mean and standard deviation of computational runtime on the ground truth foreground dataset.}
	\begin{tabular}{llrr}
		\toprule
		Setup & Method & Time [s] & Std. dev. [s]\\
		\midrule
		\multirow{4}{1.2cm}{HPC} & Multi-Level (Ours)& \textbf{2.04} & 0.296\\
		& Closed-Form \cite{levin2007closed} & 26.3 & 5.48\\
		& IndexNet \cite{lu2019indices} & 74.5 & 10.1\\
		& KNN \cite{chen2013knn} & 38.2 & 6.47\\
		\midrule
		\multirow{4}{1.2cm}{Macbook} & Multi-Level (Ours) & \textbf{1.48} & $ 0.251$\\
		& Closed-form \cite{levin2007closed} & $27.9$ &$ 7.93$\\
		& IndexNet \cite{lu2019indices} & -- & --\\
		& KNN \cite{chen2013knn} & $148.0$&  $ 56.2$\\ 
		\bottomrule
	\end{tabular}
	\label{tab:runtime}
\end{table}

We measure the runtime on two different hardware setups.
For one, we used a high-performance computer with Intel Xeon Gold 6134 CPU (3.20 GHz) and 196 GB memory. 
We also run the experiments on a MacBook Pro 2019 with Intel Core i5 (1.40 GHz) and 8 GB memory to to compare to a setup that is more realistic for everyday image processing.
To ensure comparability between the different methods, we perform all computations on the CPU on a single thread.
Table \ref{tab:runtime} compares the computational runtime of the different methods. Our method runs faster than the next best method by over an order of magnitude on both setups.

\begin{figure}
	\centering
	\includegraphics[width=0.45\textwidth]{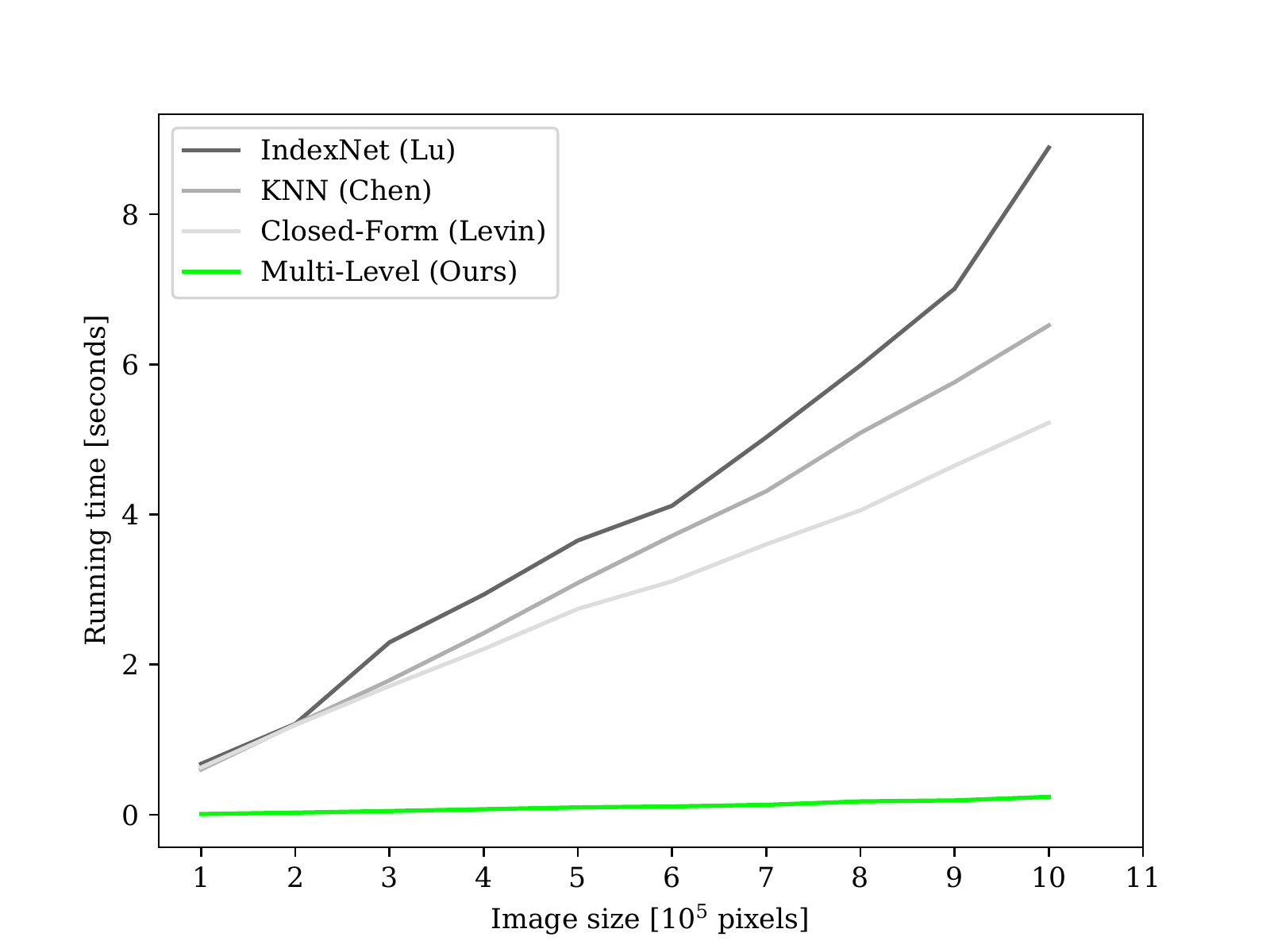} 
	\caption{Computational runtime for IndexNet, KNN, closed-form and our foreground estimation method for increasingly larger input image sizes.}
	\label{fig:runtime_plot}
\end{figure}

In addition, we compare the running time over different image sizes (Figure~\ref{fig:runtime_plot}). We can observe that, while all three methods scale roughly linearly with the image size, ours has a significantly lower constant factor.

\begin{table}
	\centering
	\caption{Maximum memory usage for each method.}
	\begin{tabular}{lrr}
		\toprule
		Method & Memory [MB] & Data Type\\
		\midrule
		Multi-Level (Ours)  & \textbf{1\,182} & 64-bit float\\
		Closed-Form \cite{levin2007closed} & $7\,781$ & 64-bit float\\
		IndexNet \cite{lu2019indices} & $91\,648$ & 32-bit float\\
		KNN \cite{chen2013knn} & $7\,850$ & 64-bit float\\
		\bottomrule
	\end{tabular}
	\label{tab:memory}	
\end{table}

Table~\ref{tab:memory} shows the memory usage for different methods. 
The IndexNet model requires most memory by far, even though its underlying data type is only half as large as that of the other methods. Therefore, this method can not be evaluated on high resolution images on the second setup. The closed-form and the KNN approach still require several gigabytes, but significantly less memory.
Finally, our multi-level approach is even more frugal in memory usage, requiring less than a sixth of the memory compared to the next best method.

\section{Conclusion}
Our proposed multi-level approach clearly outperforms all existing approaches in terms of computational runtime and memory requirements on different hardware setups while being competitive with more computationally expensive methods with respect to the quality of the estimated foreground. Additionally, our method is robust to inaccurate alpha matte estimates. This is a useful property because, for many applications, ground truth alpha mattes are not available. In this case, our method often outperforms other methods with respect to various error measures.

We have shown that our approach scales excellently with the input image size, which allows estimating the foreground of megapixel images on consumer hardware in reasonable time.

Implementations of multi-level foreground estimation for both CPU and GPU are available in the open source PyMatting library \cite{germer2020pymatting}.

{\small
	\bibliographystyle{IEEEtran}
	\bibliography{foreground}
}

\end{document}